\documentclass[letterpaper]{article} 
\usepackage{aaai25}  
\usepackage{times}  
\usepackage{helvet}  
\usepackage{courier}  
\usepackage[hyphens]{url}  
\usepackage{graphicx} 
\usepackage{natbib}  
\usepackage{caption} 
\usepackage{algorithm}
\usepackage{algorithmic}
\usepackage{amssymb}
\usepackage{mathrsfs}
\usepackage{amsmath}
\usepackage{booktabs}
\usepackage{multirow}
\usepackage[table]{xcolor}

\usepackage{newfloat}
\usepackage{listings}
\DeclareCaptionStyle{ruled}{labelfont=normalfont,labelsep=colon,strut=off} 
\floatstyle{ruled}
\newfloat{listing}{tb}{lst}{}
\floatname{listing}{Listing}
\title{Rethinking Video Segmentation with Masked Video Consistency: \\Did the Model Learn as Intended?}
\author{
    Chen Liang\equalcontrib\textsuperscript{\rm 1}\textsuperscript{\rm 3},
    Qiang Guo\equalcontrib\textsuperscript{\rm 2},
    Xiaochao Qu\textsuperscript{\rm 2},
    Luoqi Liu\textsuperscript{\rm 2},
    Ting Liu\textsuperscript{\rm 2}
}
\affiliations{
    \textsuperscript{\rm 1}Chinese Academy of Sciences, Institute of Automation \\
    \textsuperscript{\rm 2}MT Lab, Meitu Inc. \\
    \textsuperscript{\rm 3}University of Chinese Academy of Sciences \\
    liangchen2022@ia.ac.cn, \{gq5, qxc, llq5, lt\}@meitu.com
}

\usepackage{bibentry}

\begin{document}

\maketitle

\begin{abstract}

Video segmentation aims at partitioning video sequences into meaningful segments based on objects or regions of interest within frames. Current video segmentation models are often derived from image segmentation techniques, which struggle to cope with small-scale or class-imbalanced video datasets. This leads to inconsistent segmentation results across frames. To address these issues, we propose a training strategy Masked Video Consistency, which enhances spatial and temporal feature aggregation. MVC introduces a training strategy that randomly masks image patches, compelling the network to predict the entire semantic segmentation, thus improving contextual information integration. Additionally, we introduce Object Masked Attention (OMA) to optimize the cross-attention mechanism by reducing the impact of irrelevant queries, thereby enhancing temporal modeling capabilities. Our approach, integrated into the latest decoupled universal video segmentation framework, achieves state-of-the-art performance across five datasets for three video segmentation tasks, demonstrating significant improvements over previous methods without increasing model parameters.

\end{abstract}

%

\section{Introduction}
\label{sec:intro}

Video segmentation is a fundamental task in computer vision that involves partitioning a video sequence into meaningful segments, typically based on the objects or regions of interest within the frames \cite{kim2020video, qi2022occluded, yang2023panoptic}. This process is crucial for various applications, including video editing \cite{koyuncuouglu2023segmentation}, autonomous driving \cite{zhang2016instance}, action recognition \cite{moniruzzaman2021human}, and augmented reality \cite{alhaija2017augmented}. Different from image segmentation methods \cite{cheng2022masked, xie2021segformer}, modern video segmentation models \cite{huang2022minvis, ying2023ctvis} leverage object queries to establish association across frames, leading to satisfying performance. Furthermore, decoupled video segmentation framework \cite{zhang2023dvis, zhang2023dvis++} decouples the segmentation process into seperate components for spatial feature extraction, object association, and temporal feature aggregation, allowing each part to be optimized independently and improving overall segmentation accuracy and consistency. 

However, existing video segmentation methods still encounter significant challenges. Firstly, these models typically build on image segmentation models, which struggle to comprehend the general shape, texture, and contour of objects in small-scale or class-imbalanced datasets, often leading to the segmentation of a single object into multiple regions. Secondly, the segmentation results across the temporal dimension lack consistency, resulting in unstable segmentation of the same object across adjacent frames. Lastly, in query-based models, an excessive number of queries leads to slow convergence \cite{cheng2022masked, li2023dropkey}, as utilizing cross-attention mechanisms to locate target segmentation regions in both spatial and temporal dimensions requires numerous training epochs. These issues hinder the robustness of the model in challenging real-world scenarios.

\begin{figure}[!t]
\begin{center}
\includegraphics[width=0.9\linewidth]{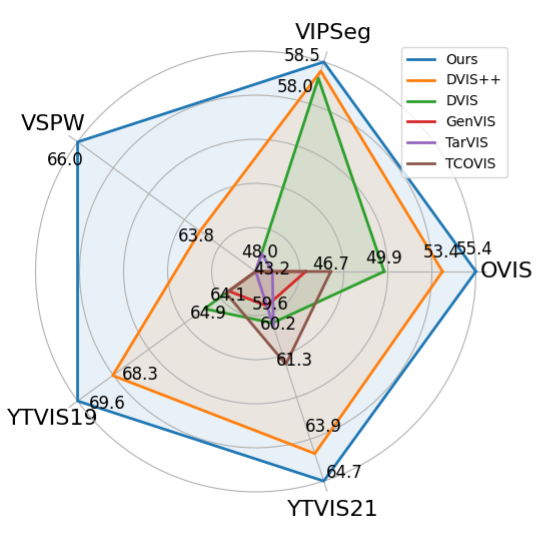}
\caption{Our method outperforms other methods in VPS, VSS, and VIS tasks, validated on five different datasets. All improvements have not added any additional model parameters. } 
\label{fig:radar}
\end{center}
\end{figure}

Further analysis reveals that video segmentation datasets contain a substantial amount of redundant information, with minimal differences between frames. This redundancy can cause models to overfit to specific shapes and textures in the training set. Additionally, these models often lack the capability to model real-world physical properties, resulting in poor performance in recognizing occluded objects. The presence of excessive background information introduces noise in the segmentation process, sometimes leading to high similarity between background queries and non-background queries. To address these issues, \textbf{is it possible to introduce additional auxiliary segmentation cues without increasing the training data?}

Indeed, it is possible. Given that segmentation models focus on object shapes and the boundaries between different objects, we propose \textbf{M}asked \textbf{V}ideo \textbf{C}onsistency (MVC) to provide additional cues meaningful for all video segmentation tasks. Specifically, we introduce a novel training strategy that randomly masks out a selection of image patches and trains the network to predict the semantic segmentation result of the entire image, including the masked-out parts. This approach helps the network aggregate contextual information across both spatial and temporal scales. For the spatial feature extraction and object association modules in the decoupled video segmentation framework, we design different masking strategies tailored for spatial and temporal dimensions, providing the model with additional and challenging training tasks. Furthermore, we propose \textbf{O}bject \textbf{M}asked \textbf{A}ttention (OMA) to improve the existing temporal feature aggregation module. By reducing the influence of irrelevant queries in the cross-attention mechanism, we achieve stronger temporal modeling capabilities for objects. 

To validate the effectiveness of our proposed Masked Video Consistency (MVC) and Object Masked Attention (OMA), we conducted experiments based on the latest decoupled universal video segmentation framework~\cite{zhang2023dvis++}. We evaluated our approach across five different datasets for three video segmentation tasks. As illustrated in Fig. \ref{fig:radar}, our method achieved performance improvements in all tasks without increasing any of the model parameters.

In summary, our contribution are as follows:

\begin{itemize}
    \item To address the challenges of object shape, texture, and contour comprehension in small-scale or class-imbalanced datasets, as well as the inconsistency of segmentation results across the temporal dimension, we introduce \textbf{M}asked \textbf{V}ideo \textbf{C}onsistency (MVC), which masks out random image patches and trains the network to predict the semantic segmentation result of the entire image. This strategy enhances the network's ability to aggregate contextual information across spatial and temporal scales.
    
    \item To further enhance the feature aggregation ability of queries in the entire temporal-spatial space in query-based models. We propose \textbf{O}bject \textbf{M}asked \textbf{A}ttention (OMA). OMA reduces the influence of irrelevant queries in the cross-attention mechanism, allowing for more efficient training and better handling of temporal dependencies in video segmentation tasks.
    
    \item We validate the effectiveness of our proposed MVC and OMA approaches by evaluating across five different datasets for three video segmentation tasks. The results demonstrate performance improvements in all tasks without increasing the number of model parameters. 
\end{itemize}

\begin{figure*}[t]
\begin{center}
\includegraphics[width=0.8\linewidth]{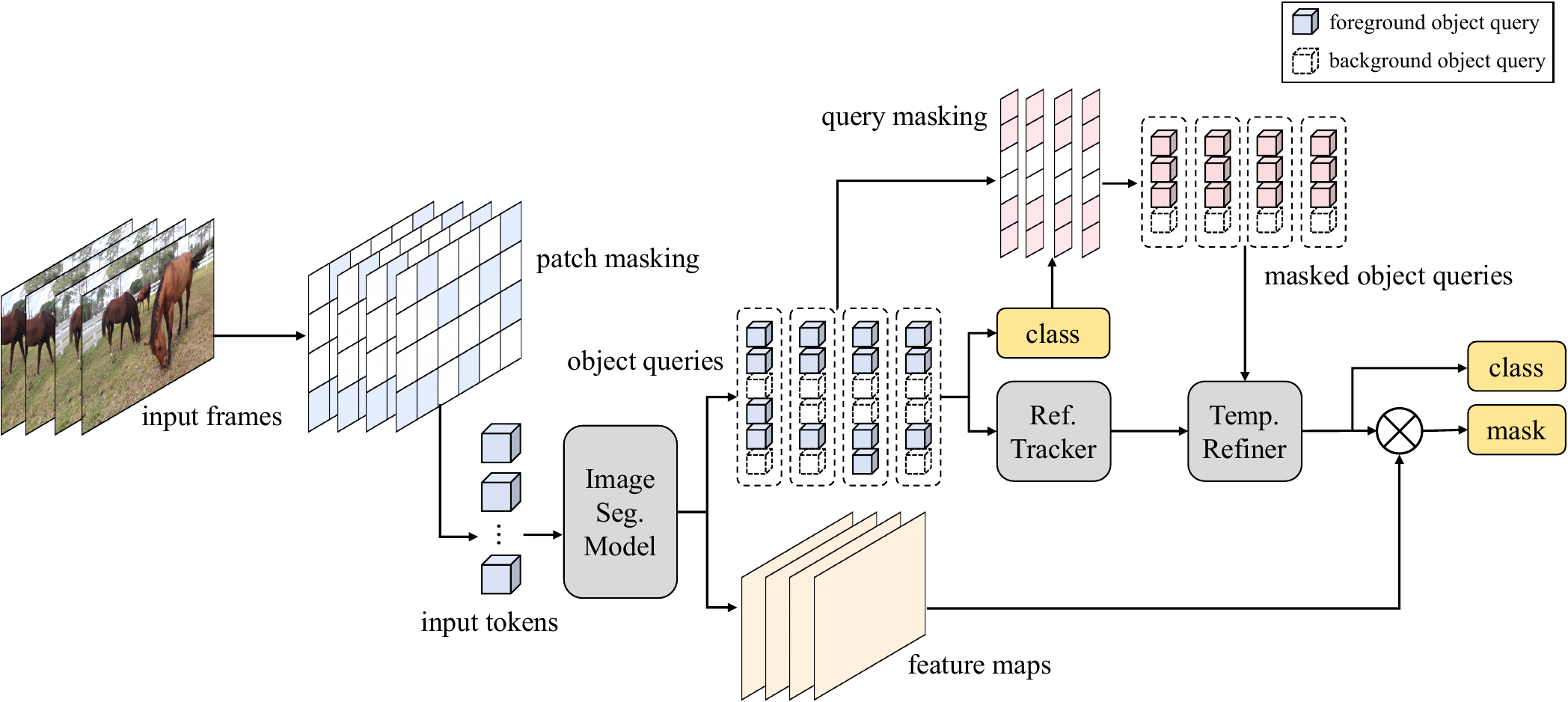}
\caption{The overall decoupled video segmentation framework equipped with Masked Video Consistency (MVC) and Object Masked Attention (OMA). Patch masking and query masking use colored blocks to indicate elements retained after element-wise masking, while uncolored areas indicate removal. In the object query diagrams, the colored cubes represent foreground object queries, and the uncolored dashed cubes represent background object queries.} 
\label{fig:framework}
\end{center}
\end{figure*}

\section{Related Work}
\label{sec:related}

\subsection{Universal Video Segmentation}

With the evolution of deep learning model architecture, more and more tasks tend to use generic models to efficiently solve various subtasks. Based on universal image segmentation methods such as \cite{cheng2021per, cheng2022masked, wang2021max, zhang2021k, yu2022k}, universal video segmentation methods have achieved comparable or even higher performance compared to specialized methods. Among these, Video K-Net \cite{li2022video} achieves video segmentation by tracking and associating object kernels, while TubeFormer \cite{kim2022tubeformer}, Tube-Link \cite{li2023tube}, and TarVIS \cite{athar2023tarvis} perform video segmentation by directly segmenting objects within clips and associating segmentation results between clips through overlapping frames or similarity clues.
DVIS \cite{zhang2023dvis} and DVIS++ \cite{zhang2023dvis++} employ a trainable referring tracker for more elegant and efficient object association. Additionally, they effectively model the spatio-temporal representation of objects throughout the entire video using a temporal refiner. However, previous work completely set the training objective of the model as the segmentation task itself. In fact, the dataset provides a large number of semantic clues beyond pixel categories, and mining these additional information can significantly broaden and enhance the model's ability boundaries.

\subsection{Masked Image / Video Modeling}

In natural language processing, the use of masked tokens has become a powerful self-supervised pretraining technique \cite{brown2020language, devlin2018bert}, where the general approach involves predicting the withheld tokens within a partially masked input sequence. Similarly, in the field of computer vision, several works have explored the concept of masking portions of an image and then extracting features from the masked regions, such as VAE features \cite{bao2021beit, dong2023peco, li2022mc}, HOG features \cite{wei2022masked}, or direct color information \cite{he2022masked, xie2022simmim}. He et al. introduced the MAE \cite{he2022masked} self-supervised learning model, which employs random patch masking. The effectiveness of other strategies, including block-wise masking \cite{bao2021beit} and attention-guided masking \cite{kakogeorgiou2022hide, li2021mst}, has also been validated. 

MIC \cite{hoyer2023mic} leverages masked images to learn contextual relations for domain adaptation, further extending masked image modeling to supervised image learning tasks. Zhang et al. \cite{zhang2023temporality} extended the MIC mechanism to the domain adaptive video segmentation task, incorporating optical flow to facilitate the learning of spatial context relations in video sequences. Unlike previous works, we directly incorporate masked video modeling in fully supervised tasks, thoroughly exploring its effectiveness. Additionally, our approach not only applies targeted masking to input images but also implements specialized masking designs within the network itself, further enhancing model capabilities.

\section{Method}
\label{sec:method}

\subsection{Decoupled Video Segmentation Frameworks}

Decoupled video segmentation frameworks \cite{zhang2023dvis, zhang2023dvis++} break down the video segmentation task into three decoupled and progressively advancing sub-tasks. They first extract object representations from each frame using a query-based image segmentation method, resulting in object representations \(Q_{Seg}\), confidence scores \(S\), and segmentation masks \(M\). In the second stage, the model maintains object consistency and tracking across frames by updating object representations \(Q_{RT}\) using the current frame's \(Q_{Seg}\) and the previous frame's reference. Finally, the third stage combines information from multiple frames to generate a spatio-temporal representation \(Q_{TR}\), which captures motion and temporal coherence for each object. A more detailed introduction to the principles of each stage can be found in App. \ref{sec:decoupled}.




\subsection{Masked Video Consistency (MVC)}

To accurately identify an object or region, a model typically needs to utilize clues from different parts of a video, including the local information of the same semantic category region or the contextual information of the same frame, as well as the semantic features from other video frames which constitute a crucial temporal context.

\begin{figure}[!t]
\begin{center}
\includegraphics[width=1\linewidth]{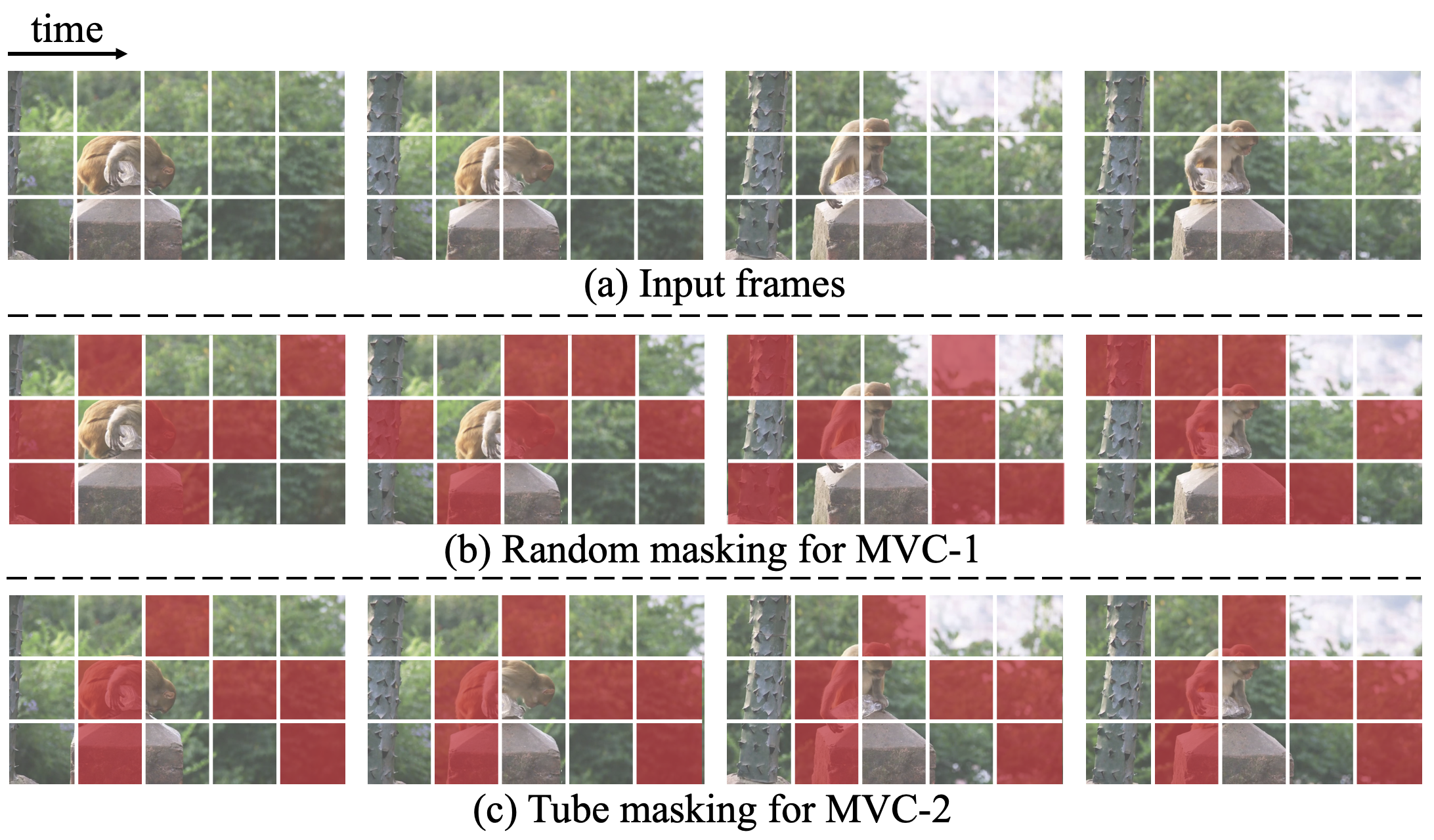}
\caption{Visualization of different masking strategy for MVC-1 and MVC-2. The red blocks represent the patches of the image that are masked out.} 
\label{fig:mvc}
\end{center}
\end{figure}

Many network architectures \cite{huang2022minvis, ying2023ctvis, zhang2023dvis} have the capability to integrate local and contextual information into their features. However, they struggle to effectively enforce the learning of useful context clues. Our experiments have shown that while current methods \cite{zhang2023dvis, zhang2023dvis++} model inter-frame relationships within a unified video through the Tracker and Refiner, it still exhibits instability in the final segmentation results. To inject more contextual clues into the model, we propose the plug-and-play Masked Video Consistency (MVC) training strategy. Fig. \ref{fig:framework} demonstrates the overall framework equipped with MVC. Specifically, for the objectives of the 1st and 2nd stages, we have designed distinct MVC training strategies tailored to each task, as shown in Fig. \ref{fig:mvc}.

\begin{figure}[!t]
\begin{center}
\includegraphics[width=0.85\linewidth]{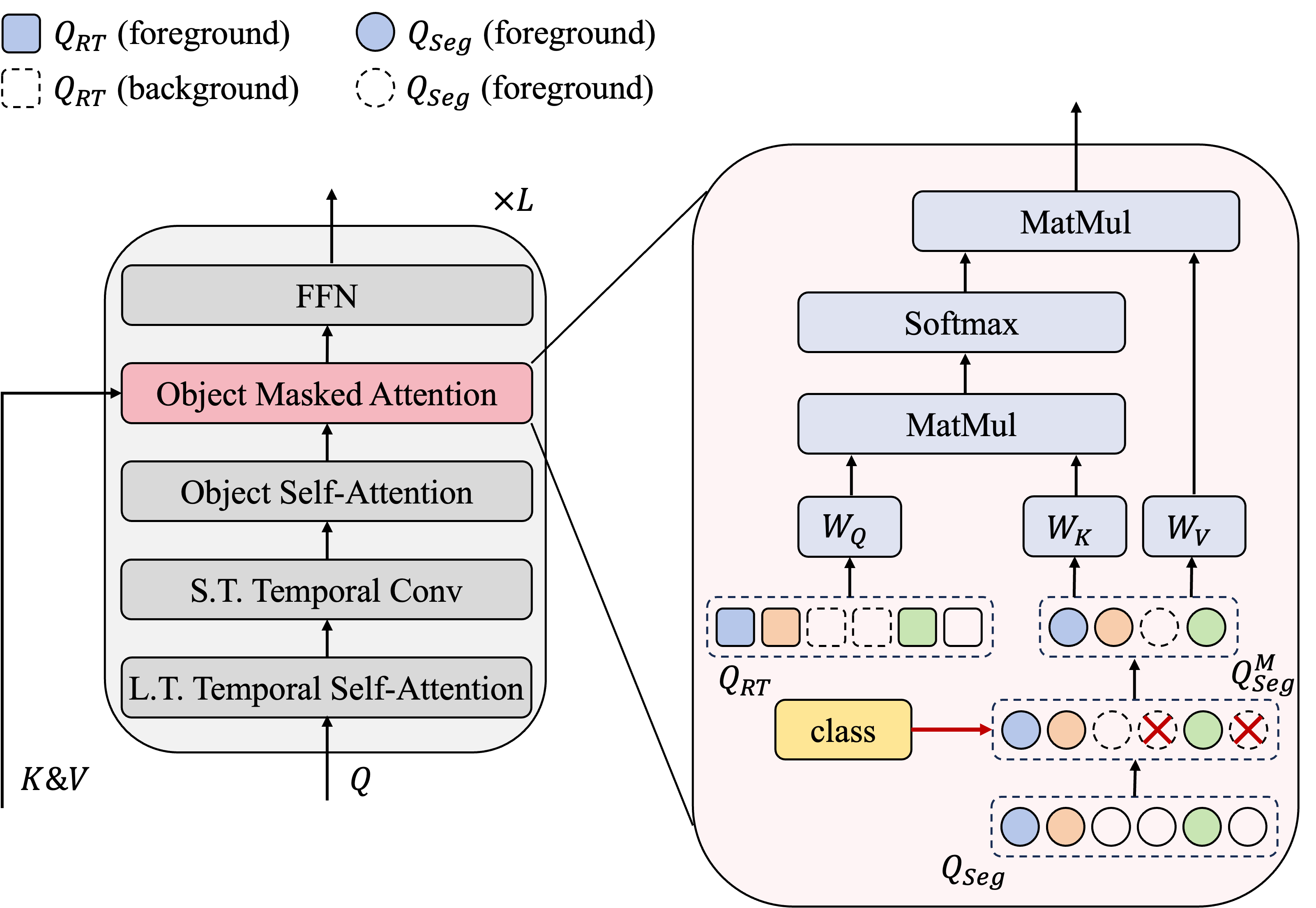}
\caption{The mechanism of Object Masked Attention. $Q_{Seg}$ represents the query output of the Segmenter. $Q_{RT}$ represents the query output of the Tracker. Different colors indicate different objects, while dashed shapes represents background object queries. } 
\label{fig:oma}
\end{center}
\end{figure}

\paragraph{MVC for Query-based Image Segmentation Method}

For each image, a patch Mask $\mathcal{M}$ is obtained by

\begin{equation}
\label{eq:mask_gen}
\mathcal{M}_{ij} = \left[ v > r \right] \quad \text{with} \quad v \sim \mathcal{U}(0, 1)
\end{equation}

where $[\cdot]$ denotes the Iverson bracket, $b$ is the patch size, $r$ is the mask ratio, $i$ and $j$ represent the positions of image blocks. The masked image is obtained by element-wise multiplication of mask and image

\begin{equation}
    I^M = \mathcal{M} \odot I
\end{equation}

The final results are inferred from the images that were removed from a portion of the mask

\begin{equation}
    \hat{Q}_{Seg}, \hat{S}, \hat{M} = \mathcal{S}(I^M)
\end{equation}

Under stage 1, given the loss $\mathcal{L}$ that defined according to the baseline model, the loss function obtained for model training is as follows

\begin{equation}
\label{eq:mask_loss}
    \underset{\theta}{\text{min}} \frac{1}{T} \sum_{t=1}^{T} \frac{1}{N} \sum_{k=1}^{N} \left( \mathbb{I}_{k, \tau} \mathcal{L}_k + (1 - \mathbb{I}_{k, \tau}) \lambda \mathcal{L}_k^M \right)
\end{equation}

where $\mathbb{I}_{k, \tau}$ is a Bernoulli random indicator function with a probability $\tau$ of being $1$ and probability $1 - \tau$ of being $0$. $\tau$ represents the probability of adding image-level masks as input. 

\paragraph{MVC for Association component}

For each video sequence consisting of \( T \) frames, a patch mask \(\mathcal{M}\) is obtained by Eq. \ref{eq:mask_gen}. Due to the temporal redundancy and temporal correlation characteristics of video sequences \cite{tong2022videomae, wang2023videomae}, we use tube masking to avoid random masking being too simple and ineffective in video tasks, while encouraging the network to fully utilize context for semantic reconstruction. In other words, the same mask \(\mathcal{M}\) is applied to all frames of the video sequence. The masked frames are obtained by element-wise multiplication of the mask and each frame of the video. The difference is visualized in Fig. \ref{fig:mvc}. 

\begin{equation}
    I_t^M = \mathcal{M} \odot I_t \quad \text{for} \; t = 1, 2, \ldots, T
\end{equation}

The remaining parts of the model computation and parameters are consistent with the original training strategy. The loss function is the same as in Eq. \ref{eq:mask_loss}.

\subsection{Object Masked Attention}

In the third stage, queries from different frames interact through cross-attention. For a unified segmentation framework across different tasks, the number of queries representing each frame is fixed. However, the number of segmented regions varies significantly across different videos, and even within the same video. The target segmented regions can differ greatly depending on the task. An excessive number of object representation queries can bring difficulty to convergence, and highly similar background queries can negatively impact the final segmentation results~\cite{cheng2022masked, li2023dropkey}. We argue that not all background queries need to participate in training. For a single frame within a video, using all foreground queries and a subset of background queries can extract sufficiently high-quality feature information.

The cross-attention interactions between different frames implemented in the Refiner can be represented by the following equation:

\begin{equation}
Q^{(l+1)}_{RT} = \text{softmax}\left( \frac{Q^{(l)} {K^{(l)}}^T}{\sqrt{C}} \right) \cdot V^{(l)} + Q^{(l)}_{RT}
\end{equation}

Here, $l$ is the index of Transformer block. $Q_{l} \in \mathbb{R}^{N \times C}$ is the object representation obtained by updating historical frame information at the $l^{\text{th}}$ block, which comes from the $2^{\text{nd}}$ stage. The $Q^{(l)}$ is under transformation $f_Q(\cdot)$ from $Q^{(l)}_{RT}$, while the $ K^{(l)}$ and $V^{(l)}$ are under transformation $f_K(\cdot)$ and $f_V(\cdot)$ from $Q^{(l)}_{seg}$. 



\begin{table}[ht]
\centering
\resizebox{\columnwidth}{!}{
\begin{tabular}{l|l|cccc}
\toprule \toprule
\textbf{Method} & \textbf{Backbone} & \textbf{VPQ} & \textbf{VPQ$^{Th}$} & \textbf{VPQ$^{St}$} & \textbf{STQ} \\ \midrule
VPSNet \cite{kim2020video} & ResNet-50 & 14.0 & 14.0 & 14.2 & 20.8 \\
VPSNet-SiamTrack \cite{woo2021learning} & ResNet-50 & 17.2 & 17.3 & 17.3 & 21.1 \\
VIP-Deeplab \cite{qiao2021vip} & ResNet-50 & 16.0 & 12.3 & 18.2 & 22.0 \\
Clip-PanoFCN \cite{miao2022large} & ResNet-50 & 22.9 & 25.0 & 20.8 & 31.5 \\
Video K-Net \cite{li2022video} & ResNet-50 & 26.1 & - & - & 31.5 \\
TarVIS \cite{athar2023tarvis} & ResNet-50 & 33.5 & 39.2 & 28.5 & 43.1 \\
Tube-Link \cite{li2023tube} & ResNet-50 & 39.2 & - & - & 39.5 \\
Video-kMax \cite{shin2024video} & ResNet-50 & 38.2 & - & - & 39.9 \\
DVIS(online) \cite{zhang2023dvis} & ResNet-50 & 39.4 & 38.6 & 40.1 & 36.3 \\
DVIS(offline) \cite{zhang2023dvis} & ResNet-50 & 43.2 & 43.6 & 42.8 & \textbf{42.8} \\
\rowcolor{gray!30}
Ours(online) & ResNet-50 & 42.0 & 41.6 & 42.3 & 38.1 \\
\rowcolor{gray!30}
Ours(offline) & ResNet-50 & \textbf{44.1} & \textbf{44.1} & \textbf{44.2} & 42.7 \\
\midrule
TarVIS \cite{athar2023tarvis} & Swin-L & 48.0 & 58.2 & 39.0 & 52.9 \\
DVIS(online) \cite{zhang2023dvis} & Swin-L & 54.7 & 54.8 & 54.6 & 47.7 \\
DVIS(offline) \cite{zhang2023dvis} & Swin-L & 57.6 & 59.9 & 55.5 & 55.3 \\
DVIS++(online) \cite{zhang2023dvis++} & ViT-L & 56.0 & 58.0 & 54.3 & 49.8 \\
DVIS++(offline) \cite{zhang2023dvis++} & ViT-L & 58.0 & \textbf{61.2} & 55.2 & \textbf{56.0} \\
\rowcolor{gray!30}
Ours(online) & ViT-L & 57.3 & 59.5 & 55.3 & 50.6 \\
\rowcolor{gray!30}
Ours(offline) & ViT-L & \textbf{58.5} & 61.1 & \textbf{56.2} & 54.8 \\ 
\bottomrule \bottomrule
\end{tabular}}
\caption{Comparison of different methods on the VIPSeg validation set. VPQ$^{Th}$ and VPQ$^{St}$ refer to the VPQ (Video Panoptic Quality) on the "thing" objects and the "stuff" objects, respectively. The best results are highlighted in bold.}
\label{tab:vipseg}
\end{table}

To ensure that the model focuses more on the information and feature updates of foreground objects, we generate a mask applied to the attention map. This mask randomly drops a portion of background object queries with a certain probability, resulting in the Object Masked Attention mechanism.

\begin{equation}
\hat{Q}^{(l+1)}_{RT} = \text{softmax}\left(\mathcal{M}^{(l)} + \frac{Q^{(l)} {K^{(l)}}^T}{\sqrt{C}} \right) \cdot V^{(l)} + Q^{(l)}_{RT}
\end{equation}

The attention mask $\mathcal{M}^{(l)}$ is defined as follows

\begin{equation}
\mathcal{M}_{i} = 
\begin{cases}
0 & \text{if the $i$-th object is foreground} \\
-\infty \cdot \mathbb{I}_{i, \gamma} & \text{if the $i$-th object is background}
\end{cases}
\end{equation}

where $ \mathbb{I}_{i, \gamma} \sim \text{Bernoulli}(\gamma) $ is a Bernoulli random variable with probability $\gamma$ of being $1$ and probability $1 - \gamma$ of being $0$. Whether an object is foreground or background is determined by the confidence scores $S$, with a threshold of $0.6$ for the empty class.

\section{Experiments}
\label{sec:exp}


\paragraph{Implementation Details}

We emply the AdamW optimizer \cite{loshchilov2017decoupled} with an initial learning rate of $1e-4$ and a weight dacay of $5e-2$ for training. All relevant parameters of the model architecture are consistent with DVIS++ \cite{zhang2023dvis++}. We use Mask2Former \cite{cheng2022masked} as the Segmenter to train 60k iterations for all tasks. For VSS and VPS tasks, we train 40k iterations on Tracker and Refiner respectively. For the VIS task, we trained 160K iterations on Tracker and Refiner respectively, utilize COCO pseudo videos \cite{wu2022seqformer} for joint training. All experiments are conducted on 8 NVIDIA V100s. 

\subsection{Comparison with Exsiting Methods}

\subsubsection{Performance on Video Panoptic Segmentation}

\paragraph{VIPSeg} To validate the effectiveness of our proposed strategy, we first conducted experiments on the most challenging panoptic segmentation task, which requires the model to simultaneously achieve high performance in both semantic segmentation and instance segmentation. 

We conducted our experiments on the VIPSeg \cite{miao2022large} validation set. The results are shown in Tab. \ref{tab:vipseg}. VIPSeg is a comprehensive dataset designed for evaluating panoptic segmentation models, encompassing a wide variety of scenes and object categories, which makes it particularly challenging. Using ResNet-50 \cite{he2016deep} and ViT-L \cite{dosovitskiy2020image} as backbones, we achieved new state-of-the-art results in both online and offline modes. Specifically, our method outperformed the previous best method, DVIS++, by 1.3\% and 0.5\% with ViT-L in the online mode and the offline mode, seperately. These results highlight the robustness and adaptability of our proposed approach in handling complex segmentation tasks.

\subsubsection{Performance on Video Semantic Segmentation}

\begin{table}[ht]
\centering
\resizebox{\columnwidth}{!}{
\begin{tabular}{l|l|ccc}
\toprule \toprule
\textbf{Method} & \textbf{Backbone} & \textbf{mVC$_8$} & \textbf{mVC$_{16}$} & \textbf{mIOU} \\ \midrule
Mask2Former \cite{cheng2022masked} & ResNet-50 & 87.5 & 82.5 & 38.4 \\
Video-kMax \cite{shin2024video} & ResNet-50 & 86.0 & 81.4 & 44.3 \\
Tube-Link \cite{li2023tube} & ResNet-50 & 89.2 & 85.4 & 43.4 \\
MPVSS \cite{weng2024mask} & ResNet-50 & 84.1 & 77.2 & 37.5 \\
DVIS(online) \cite{zhang2023dvis} & ResNet-50 & 92.0 & 90.9 & 46.6 \\
DVIS(offline) \cite{zhang2023dvis} & ResNet-50 & 93.2 & 92.3 & 47.2 \\
\rowcolor{gray!30}
Ours(online) & ResNet-50 & 93.0 & 91.9 & 47.7 \\
\rowcolor{gray!30}
Ours(offline) & ResNet-50 & \textbf{93.3} & \textbf{92.4} & \textbf{49.0} \\
\midrule
DeepLabv3+ \cite{chen2018encoder} & ResNet-101 & 83.5 & 78.4 & 35.7 \\
TCB \cite{miao2021vspw} & ResNet-101 & 86.9 & 82.1 & 37.5 \\
Video K-Net \cite{li2022video} & ResNet-101 & 87.2 & 82.3 & 38.0 \\
MRCFA \cite{sun2022mining} & MiT-B2 & 90.9 & 87.4 & 49.9 \\
CFFM \cite{sun2022coarse} & MiT-B5 & 90.8 & 87.1 & 49.3 \\
Video K-Net+ \cite{li2022video} & ConvNeXt-L & 90.1 & 87.8 & 57.2 \\
Video kMax \cite{shin2024video} & ConvNeXt-L & 91.8 & 88.6 & 63.6 \\
TubeFormer \cite{kim2022tubeformer} & Axial-ResNet-50 & 92.1 & 88.0 & 63.2 \\
MPVSS \cite{weng2024mask} & Swin-L & 89.6 & 85.8 & 53.9 \\
DVIS(online) \cite{zhang2023dvis} & Swin-L & 95.0 & 94.3 & 61.3 \\
DVIS(offline) \cite{zhang2023dvis} & Swin-L & 95.1 & 94.4 & 63.3 \\
DVIS++(online) \cite{zhang2023dvis++} & ViT-L & 95.0 & 94.2 & 62.8 \\
DVIS++(offline) \cite{zhang2023dvis++} & ViT-L & \textbf{95.7} & \textbf{95.1} & 63.8 \\ 
\rowcolor{gray!30}
Ours(online) & ViT-L & 95.6 & 94.9 & 64.4 \\
\rowcolor{gray!30}
Ours(offline) & ViT-L & \textbf{95.7} & \textbf{95.1} & \textbf{66.0} \\ 
\bottomrule \bottomrule
\end{tabular}}
\caption{Comparison of different methods on the VSPW validation set. mVC$_k$ means that a clip with $k$ frames is used. The best results are highlighted in bold.}
\label{tab:vspw}
\end{table}

\paragraph{VSPW} To further validate our proposed strategy, we conducted experiments on the semantic segmentation task, which focuses on accurately classifying each pixel in an image. We conducted our experiments on the VSPW \cite{miao2021vspw} validation set, as shown in Tab. \ref{tab:vspw}. 

We achieved new state-of-the-art results in both online and offline modes using ResNet-50 and ViT-L as backbones seperately. Taken DVIS++ as the baseline, our method outperformed it by 1.6\% with ViT-L in the offline mode. The improvements in the online mode were similarly significant with 2.2\% improvement on ViT-L backbone. It is worth mentioning that when using ViT-L as the backbone, the results of our online model are already better than the offline model of DVIS++. Our results demonstrate that our method surpasses all previous models without adding any additional parameters, and even reduces the network training burden. 

\subsubsection{Performance on Video Instance Segmentation}

\begin{table*}[t]
\centering
\resizebox{0.9\textwidth}{!}{%
\begin{tabular}{llccccccccc}
\toprule
\toprule
\multicolumn{1}{c|}{\multirow{2}{*}{\textbf{Method}}} & \multicolumn{1}{c|}{\multirow{2}{*}{\textbf{Backbone}}} & \multicolumn{3}{c|}{\textbf{OVIS}} & \multicolumn{3}{c|}{\textbf{YT-VIS 2019}} & \multicolumn{3}{c}{\textbf{YT-VIS 2021}} \\
\multicolumn{1}{c|}{} & \multicolumn{1}{c|}{} & \multicolumn{1}{c}{\textbf{AP}} & \multicolumn{1}{c}{\textbf{AP$_{75}$}} & \multicolumn{1}{c|}{\textbf{AR$_{10}$}} & \multicolumn{1}{c}{\textbf{AP}} & \multicolumn{1}{c}{\textbf{AP$_{75}$}} & \multicolumn{1}{c|}{\textbf{AR$_{10}$}} & \multicolumn{1}{c}{\textbf{AP}} & \multicolumn{1}{c}{\textbf{AP$_{75}$}} & \multicolumn{1}{c}{\textbf{AR$_{10}$}} \\ \midrule
\multicolumn{1}{c|}{MinVIS \cite{huang2022minvis}} & \multicolumn{1}{c|}{ResNet50} & 25.0 & 24.0 & \multicolumn{1}{c|}{29.7} & 47.4 & 52.1 & \multicolumn{1}{c|}{55.7} & 44.2 & 48.1 & 51.7 \\
\multicolumn{1}{c|}{CTVIS \cite{ying2023ctvis}} & \multicolumn{1}{c|}{ResNet50} & 35.5 & 34.9 & \multicolumn{1}{c|}{41.9} & 55.1 & 59.1 & \multicolumn{1}{c|}{63.2} & 50.1 & 54.7 & 59.5 \\
\multicolumn{1}{c|}{GenVIS \cite{heo2023generalized}} & \multicolumn{1}{c|}{ResNet50} & 35.8 & 36.2 & \multicolumn{1}{c|}{39.6} & 50.0 & 54.6 & \multicolumn{1}{c|}{59.7} & 47.1 & 51.5 & 54.7 \\
\multicolumn{1}{c|}{DVIS(Online) \cite{zhang2023dvis}} & \multicolumn{1}{c|}{ResNet50} & 30.2 & 30.5 & \multicolumn{1}{c|}{37.3} & 51.2 & 57.1 & \multicolumn{1}{c|}{59.3} & 46.4 & 49.6 & 53.5 \\
\multicolumn{1}{c|}{DVIS(Offline) \cite{zhang2023dvis}} & \multicolumn{1}{c|}{ResNet50} & 33.8 & 33.5 & \multicolumn{1}{c|}{39.5} & 52.3 & 58.2 & \multicolumn{1}{c|}{60.4} & 47.4 & 51.6 & 55.2 \\
\rowcolor{gray!30}
\multicolumn{1}{c|}{Ours(Online)} & \multicolumn{1}{c|}{ResNet50} & 35.2 & 34.9 & \multicolumn{1}{c|}{41.8} & 55.2 & 60.0 & \multicolumn{1}{c|}{63.2} & 48.7 & 51.9 & 56.3 \\
\rowcolor{gray!30}
\multicolumn{1}{c|}{Ours(Offline)} & \multicolumn{1}{c|}{ResNet50} & \textbf{40.4} & \textbf{40.1} & \multicolumn{1}{c|}{\textbf{46.9}} & \textbf{55.6} & \textbf{60.5} & \multicolumn{1}{c|}{\textbf{63.8}} & \textbf{50.1} & \textbf{55.4} & \textbf{58.3} \\ 
\midrule
\multicolumn{1}{c|}{MinVIS \cite{huang2022minvis}} & \multicolumn{1}{c|}{Swin-L} & 39.4 & 41.3 & \multicolumn{1}{c|}{43.3} & 61.6 & 68.6 & \multicolumn{1}{c|}{66.6} & 55.3 & 62.0 & 60.8 \\
\multicolumn{1}{c|}{CTVIS \cite{ying2023ctvis}} & \multicolumn{1}{c|}{Swin-L} & 46.9 & 47.5 & \multicolumn{1}{c|}{52.1} & 65.6 & 72.2 & \multicolumn{1}{c|}{70.4} & 61.2 & 68.8 & 65.8 \\
\multicolumn{1}{c|}{GenVIS \cite{heo2023generalized}} & \multicolumn{1}{c|}{Swin-L} & 45.2 & 48.4 & \multicolumn{1}{c|}{48.6} & 64.0 & 68.3 & \multicolumn{1}{c|}{69.4} & 59.6 & 65.8 & 65.0 \\
\multicolumn{1}{c|}{DVIS(Online) \cite{zhang2023dvis}} & \multicolumn{1}{c|}{Swin-L} & 45.9 & 48.3 & \multicolumn{1}{c|}{51.5} & 63.9 & 70.4 & \multicolumn{1}{c|}{69.0} & 58.7 & 66.6 & 64.6 \\
\multicolumn{1}{c|}{DVIS(Offline) \cite{zhang2023dvis}} & \multicolumn{1}{c|}{Swin-L} & 48.6 & 50.5 & \multicolumn{1}{c|}{53.8} & 64.9 & 72.7 & \multicolumn{1}{c|}{70.3} & 60.1 & 68.4 & 65.7 \\
\multicolumn{1}{c|}{DVIS++(Online) \cite{zhang2023dvis++}} & \multicolumn{1}{c|}{ViT-L} & 49.6 & 55.0 & \multicolumn{1}{c|}{54.6} & 67.7 & 75.3 & \multicolumn{1}{c|}{73.7} & 62.3 & 70.2 & 68.0 \\
\multicolumn{1}{c|}{DVIS++(Offline) \cite{zhang2023dvis++}} & \multicolumn{1}{c|}{ViT-L} & 53.4 & 58.5 & \multicolumn{1}{c|}{58.7} & 68.3 & 76.1 & \multicolumn{1}{c|}{73.4} & 63.9 & 71.5 & 69.5 \\
\rowcolor{gray!30}
\multicolumn{1}{c|}{Ours(Online)} & \multicolumn{1}{c|}{ViT-L} & 51.1 & 55.1 & \multicolumn{1}{c|}{55.6} & 68.4 & 75.9 & \multicolumn{1}{c|}{74.1} & 63.1 & 71.1 & 68.4 \\
\rowcolor{gray!30}
\multicolumn{1}{c|}{Ours(Offline)} & \multicolumn{1}{c|}{ViT-L} & \textbf{55.4} & \textbf{60.4} & \multicolumn{1}{c|}{\textbf{60.1}} & \textbf{69.6} & \textbf{76.5} & \multicolumn{1}{c|}{\textbf{75.3}} & \textbf{64.7} & \textbf{71.8} & \textbf{70.4} \\ \bottomrule \bottomrule
\end{tabular}
}
\caption{Comparison of different methods on the validation set of OVIS and YouTube-VIS 2019 \& 2021. The best results are highlighted in bold.}
\end{table*}

\paragraph{OVIS}

In our experiments on the OVIS \cite{qi2022occluded} dataset, our proposed method demonstrates significant improvements over previous state-of-the-art models, particularly in complex scenarios with occlusions and fast-moving objects. Using ResNet-50 as the backbone, our method achieves an AP of 35.2\% in the online mode and 40.4\% in the offline mode, respectively. When employing a more powerful backbone such as ViT-L, our method continues to outperform existing models significantly. Specifically, we achieve an AP of 51.1\% in the online mode and 55.4\% in the offline mode, further surpassing DVIS++ by 1.5\% and 2.0\% AP, respectively. These results underscore the robustness of our approach, especially in offline settings where our method excels at refining object representations and improving overall segmentation accuracy.

\paragraph{YouTube-VIS 2019 \& 2021}

\begin{table}[ht]
\centering
\resizebox{0.7\columnwidth}{!}{
\begin{tabular}{ccc|c|c}
\toprule \toprule
\textbf{MVC-1} & \textbf{MVC-2} & \textbf{OMA} & mIoU & VPQ$_{\text{all}}$ \\ \midrule
-- &  & \multicolumn{1}{c|}{} & 62.2 & 54.8 \\
\checkmark &  & \multicolumn{1}{c|}{} & \textbf{63.0} & \textbf{55.4} \\
-- & -- & \multicolumn{1}{c|}{} & 62.8 & 56.0 \\
-- & \checkmark & \multicolumn{1}{c|}{} & \textbf{63.6} & \textbf{56.6} \\
-- & -- & \multicolumn{1}{c|}{--} & 63.8 & 58.0 \\ 
-- & -- & \multicolumn{1}{c|}{\checkmark} & \textbf{65.0} & \textbf{58.3} \\
\bottomrule \bottomrule
\end{tabular}
}
\caption{The component ablation for the proposed MVC and OMA. "--" denotes the use of the baseline DVIS++ for training, and a checkmark signifies the incorporation of our MVC strategy or OMA mechanism for training. The mIoU and VPQ$_{\text{all}}$ represent evaluation metrics on the VSPW and VIPSeg datasets, respectively. }
\label{tab:compo}
\end{table}

The YouTube-VIS 2019 and 2021 \cite{yang2019video} datasets consist of shorter videos and simpler scenes compared to the more complex and occlusion-heavy OVIS \cite{qi2022occluded} dataset. Despite the less challenging nature of these datasets, our proposed method still demonstrates clear advantages over previous state-of-the-art models. In online mode, our method achieves an AP of 55.2\% on YouTube-VIS 2019 and 48.7\% on YouTube-VIS 2021, while achieving an AP of 55.6\% on YouTube-VIS 2019 and 50.1\% on YouTube-VIS 2021 in offline mode, using ResNet-50 as backbone and DVIS as baseline.

When adopting the ViT-L backbone, our method continues to set new benchmarks. On YouTube-VIS 2019, we achieve an AP of 68.4\% in online mode and 69.6\% in offline mode, while on YouTube-VIS 2021, we reach 64.1\% AP in online mode and 64.7\% AP in offline mode. These results not only highlight the effectiveness of our approach across different datasets but also demonstrate its superior adaptability in both online and offline settings, outperforming DVIS++ across the board.

\begin{table}[ht]
\centering
\resizebox{0.55\columnwidth}{!}{
\begin{tabular}{cccc}
\toprule \toprule
\multicolumn{1}{c|}{\textbf{Patch Size $s$}} & \multicolumn{3}{c}{\textbf{Mask Ratio $r$}} \\ \midrule
\multicolumn{4}{c}{MVC-1}                                            \\ \midrule
\multicolumn{1}{c|}{}             & \textbf{0.3}       & \textbf{0.5}       & \textbf{0.7}      \\ \midrule
\multicolumn{1}{c|}{\textbf{16}}           & 45.2          & \textbf{45.7}          & 45.6         \\
\multicolumn{1}{c|}{\textbf{32}}           & 45.0          & 45.4          & \textbf{45.7}         \\
\multicolumn{1}{c|}{\textbf{64}}           & 45.2          & \textbf{45.7}          & 45.6         \\ \midrule
\multicolumn{4}{c}{MVC-2}                                            \\ \midrule
\multicolumn{1}{c|}{}             & \textbf{0.3}       & \textbf{0.5}       & \textbf{0.7}      \\ \midrule
\multicolumn{1}{c|}{\textbf{16}}           & 47.5          & 46.3          & 47.6         \\
\multicolumn{1}{c|}{\textbf{32}}           & 47.0          & 47.6          & \textbf{47.7}         \\
\multicolumn{1}{c|}{\textbf{64}}           & 47.4          & 47.3          & 47.6         \\ 
\bottomrule \bottomrule
\end{tabular}
}
\caption{Parameter study of the patch size $s$ and the mask ratio $r$ for MVC in the 1st stage and 2nd stage. The best results are highlighted in bold. }
\label{tab:patch}
\end{table}

\begin{table}[ht]
\centering
\resizebox{0.6\columnwidth}{!}{
\begin{tabular}{cc|c}
\toprule \toprule
\textbf{Drop Ratio $d$} & \textbf{Masked Inf.} & \textbf{mIoU} \\ \midrule
\textbf{0}                      & --               & 63.8 \\ \midrule
\multirow{2}{*}{\textbf{0.2}}   & --               & 64.4 \\
                                & \checkmark       & 64.3 \\ \midrule
\multirow{2}{*}{\textbf{0.5}}   & --               & \textbf{65.0} \\
                                & \checkmark       & 64.9 \\ \midrule
\multirow{2}{*}{\textbf{0.7}}   & --               & 64.8 \\
                                & \checkmark       & 64.6 \\
\bottomrule \bottomrule
\end{tabular}
}
\caption{Parameter study of the drop ratio $d$ for OMA in the 3rd stage. Masked Inf. represents discarding keys and values with the same ratio during the testing phase. The best results are highlighted in bold. }
\label{tab:drop}
\end{table}

\subsection{Ablation Studies}

\paragraph{Component ablation}

Tab. \ref{tab:compo} presents experimental results corresponding to the three stages of the decoupled video segmentation framework from left to right. MVC-1 represents the MVC training strategy applied to the Segmenter, and MVC-2 represents the MVC training strategy applied to the Tracker. The blank cell indicates that the respective stage is not trained, "--" denotes the use of the baseline DVIS++ for training, and a checkmark signifies the incorporation of our MVC strategy or OMA mechanism for training. 

The results demonstrate improvements in both the semantic segmentation task on the VSPW dataset and the panoptic segmentation task on the VIPSeg dataset when applying each of the three components. This indicates that all our proposed contributions are meaningful and effective, while all of the parts stacking together can further enhance the effect.

\paragraph{Ablations of MVC}

Tab. \ref{tab:patch} shows the influence of the mask ratio $r$ and the mask patch size $s$ for MVC in the 1st stage and 2nd stage, validated on VSPW dataset using ResNet50 as backbone. Both stages demonstrated that a higher mask ratio leads to better performance, suggesting that forcing the model to rely more on additional cues within video frames helps capture contextual information. Furthermore, experimental results indicate that a patch size of $32 \times 32$ yielded the best results. Based on the analysis of both stages, we adopted a patch size of $s = 32$ and a mask ratio of $r = 0.7$ for training the first two stages in all subsequent experiments.

\paragraph{Ablations of OMA}

Tab. \ref{tab:drop} demonstrates the impact of the drop ratio for keys and values in Object Masked Attention on the final results. Notably, we retained all object queries predicted by the network as foreground and only applied random dropout to background object queries during training. A drop ratio $d$ of 0.5 yielded the best performance, indicating that while retaining all background object queries can negatively affect model convergence, some of them contain valuable information. In Tab. \ref{tab:drop}, "Masked Inf." represents discarding keys and values with the same ratio during the testing phase. The results suggest that retaining all queries during testing is the optimal choice, as they provide meaningful reference points for the network's attention calculations.

\paragraph{Qualitative Analysis}

\begin{figure}[!t]
\begin{center}
\includegraphics[width=1\linewidth]{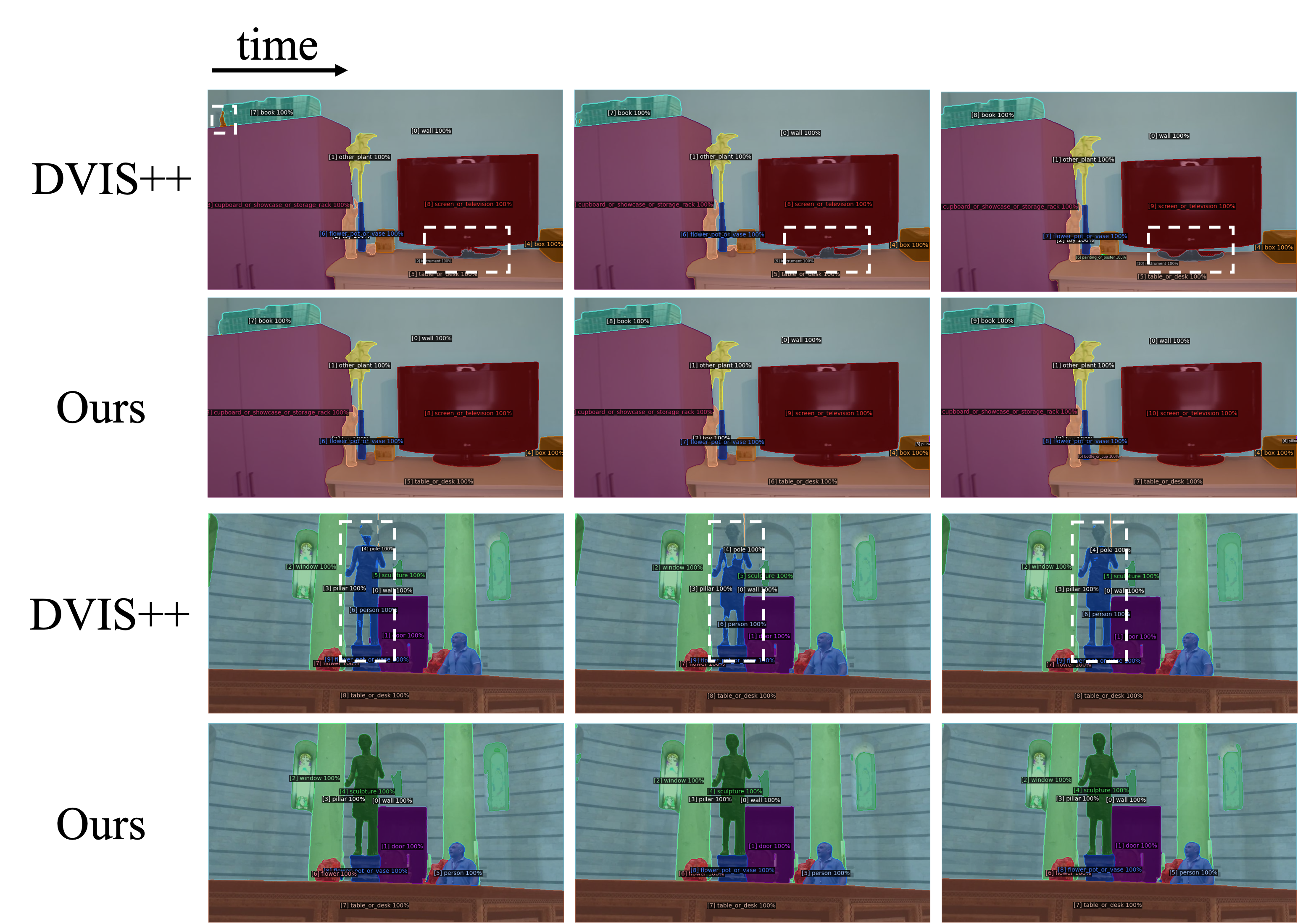}
\caption{Qualitative comparison for baseline model and our model. The white dashed box indicates the segmentation problem in the baseline model. } 
\label{fig:comp}
\end{center}
\end{figure}

\begin{figure}[!t]
\begin{center}
\includegraphics[width=1\linewidth]{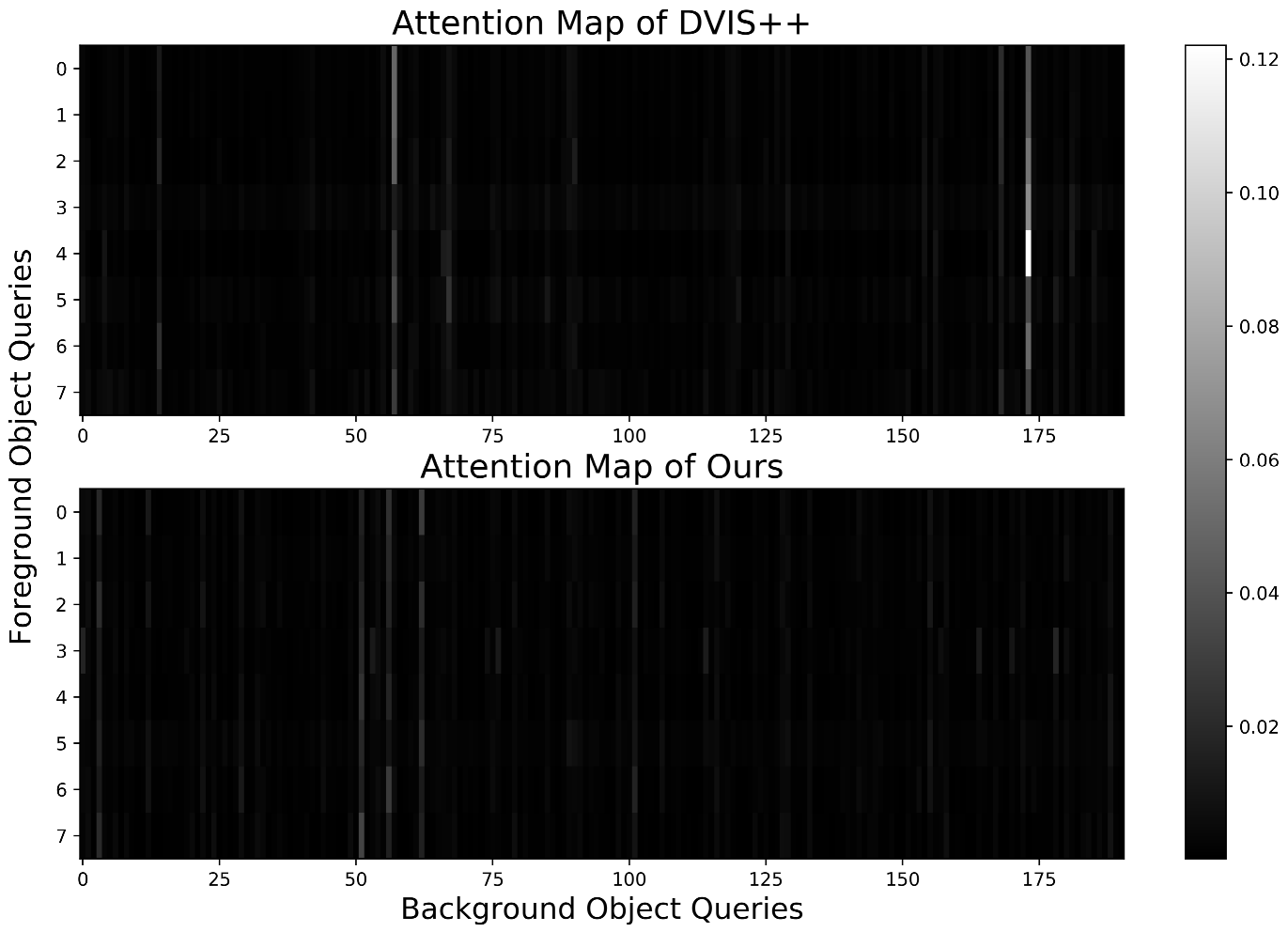}
\caption{Visualization of the cross-attention map during inferencing. Brighter pixel values represent higher attention weight. } 
\label{fig:attn}
\end{center}
\end{figure}

Fig. \ref{fig:comp} shows the comparison of segmentation results between our model and the baseline model DVIS++ on the VSPW dataset, displaying three consecutive frames from two videos. We highlighted the segmentation problem of the baseline model using a white dashed box. Specifically, our proposed method optimizes the spatial and temporal consistency of the segmentation model through the training strategy of MVC and OMA mechanism, which can better utilize the information exchange of consecutive multiple frames. More visualization results can be found in App. \ref{sec:add_vis}. 

To further demonstrate the impact of our proposed Object Masked Attention (OMA) on the model's attention learning, we partially visualized the cross-attention maps during the testing phase. Specifically, in Fig. \ref{fig:attn}, we present the attention weights of foreground object queries on background object queries within the same frame. Compared to DVIS++ without OMA, our approach concentrates more attention weights on foreground object queries while minimizing reliance on specific background object queries. This design enables the model to learn to extract more distinctive and background-independent features, especially in more complex and challenging scenarios.

\section{Conclusion}

Focusing on the key challenges in video segmentation, we proposed MVC and OMA as novel strategies to to enhance spatial and temporal feature aggregation, thereby improving the overall accuracy and consistency of segmentation results across frames. Through extensive experimentation across five datasets and three distinct video segmentation tasks, we demonstrated that our approach not only outperforms existing methods but also achieves these improvements without increasing model parameters. The introduction of MVC and OMA provides additional auxiliary cues that allow the model to better utilize contextual information and reduce the impact of irrelevant queries during the segmentation process. Meanwhile, our method demonstrates that self-supervised training strategies have additional value beyond providing pre-training on supervised tasks, thereby providing direction for subsequent exploration.

\bibliography{aaai25}

\newpage

\appendix
\section{Datasets}
\label{sec:datasets}

\paragraph{VSPW} VSPW \cite{miao2021vspw} is a dataset with the target of advancing the scene parsing task for images to videos. VSPW contains 3,536 videos, including 251,633 frames from 124 categories. Each video contains a well-trimmed long-temporal shot, lasting around 5 seconds on average. 

\paragraph{VIPSeg} VIPSeg \cite{miao2022large} is a large-scale dataset designed for video panoptic segmentation, featuring 3,536 videos and 84,750 frames annotated with pixel-level panoptic labels. The dataset covers 124 categories, including 58 "thing" classes and 66 "stuff" classes, and is divided into 2,806 videos for training, 343 for validation, and 387 for testing. VIPSeg captures a diverse array of real-world scenarios, making it a robust benchmark for evaluating segmentation models. 

\paragraph{OVIS} OVIS \cite{qi2022occluded} is a highly challenging dataset specifically designed for video instance segmentation tasks, particularly focusing on scenarios with significant object occlusions. The dataset consists of 901 videos, with a total of 5,223 unique object instances across 25 object categories. OVIS is known for its lengthy video sequences, averaging around 500 frames per video, which closely resemble real-world conditions where objects frequently occlude each other. This makes OVIS a critical benchmark for evaluating the robustness of segmentation models in complex and dynamic environments. The dataset is divided into 607 videos for training, 140 for validation, and 154 for testing.

\paragraph{Youtube-VIS 2019} YouTube-VIS 2019 \cite{yang2019video} is one of the first large-scale benchmarks designed for video instance segmentation, consisting of 2,883 high-resolution YouTube video clips. The dataset spans 40 object categories and features over 5,000 unique instances, providing annotations for both object segmentation and tracking. Each video clip averages around 5.6 seconds, making it well-suited for evaluating models in relatively short and simple scenes. The dataset is divided into 2,238 videos for training, 302 for validation, and 343 for testing.

\paragraph{Youtube-VIS 2021} YouTube-VIS 2021 \cite{yang2019video} is an extension of the YouTube-VIS 2019 dataset, designed to further challenge video instance segmentation models. It comprises 3,859 video clips, covering the same 40 object categories as the 2019 version but with more diverse and complex scenes. The dataset includes more than 8,000 unique object instances, with each video clip averaging around 5.5 seconds. Like its predecessor, YouTube-VIS 2021 provides instance-level annotations for segmentation and object tracking, which divided into 2,985 videos for training, 421 for validation, and 453 for testing. 

\section{Decoupled Video Segmentation Frameworks}
\label{sec:decoupled}

Decoupled video segmentation frameworks \cite{zhang2023dvis, zhang2023dvis++} firstly extract object representations from each single frame using a query-based image segmentation method. In this stage, an input image $I \in \mathbb{R}^{H \times W \times 3}$ is processed to get the object representations $ Q_{Seg} \in \mathbb{R}^{N \times C} $, confidence scores $ S \in \mathbb{R}^{N \times |C|} $, and segmentation masks $M \in \mathbb{R}^{N \times H \times W}$.

\begin{equation}
    Q_{Seg}, S, M = \mathcal{S}(I)
\end{equation}

where $N$ represents the number of objects, $C$ represents the featue channels, $|C|$ represents the category size, and $H$ and $W$ represent the height and width of the image, respectively. 

In the second stage, the model ensures that object are consistently identified and tracked across frames. This stage leverages object queries to maintain associations between objects in consecutive frames. For each frame, taking the current object representation $Q_{Seg}^{T}$ and the reference ${Ref}^{T-1}$ as input, the model outputs the updated objecty representation $Q_{RT}^T$ and a new reference ${Ref}^T$ for next frame. 

\begin{equation}
    Q_{RT}^T, {Ref}^T = \mathcal{T}({Ref}^{T-1}, Q_{Seg}^T).
\end{equation}

The third stage aims to combine information across multiple frames to capture motion and temporal coherence. After the above operations, $Q_{Seg}$ is the object representation directly output by the image segmentation model, $Q_{RT}$ is the object representation obtained by updating historical frame information. Based on these two feature information, each object representation can be further optimized to ultimately obtain spatio-temporal representation $Q_{TR} \in \mathbb{R}^{N \times T \times C}$. 

\begin{equation}
    Q_{TR} = \mathcal{R}(Q_{RT}, Q_{Seg}).
\end{equation}

\section{More Qualitative Results}
\label{sec:add_vis}

\begin{figure*}[t]
\begin{center}
\includegraphics[width=0.8\linewidth]{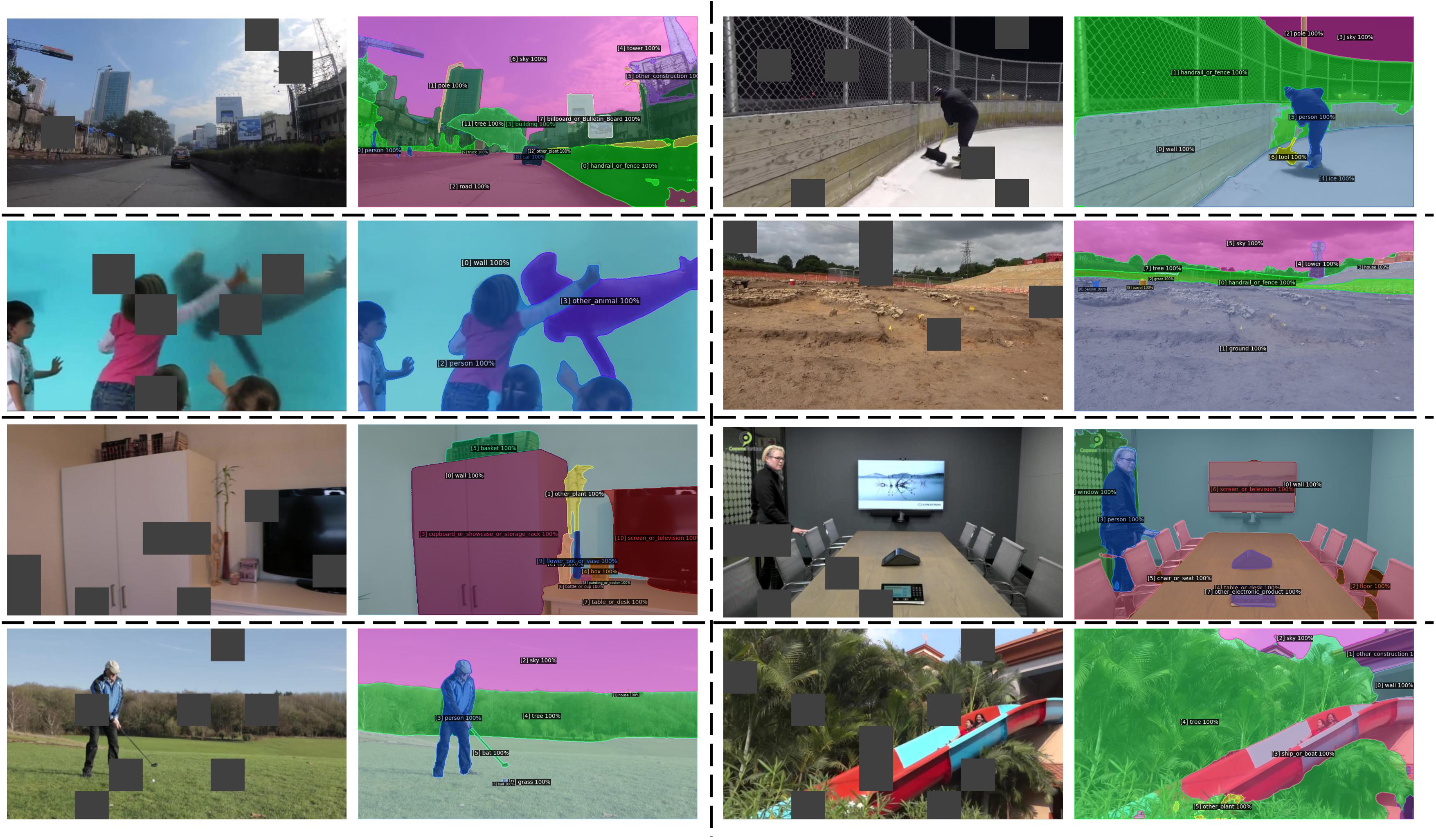}
\caption{Visualization the segmentation effect of the optimized model using MVC learning strategy. The gray patch represents that the corresponding part in the image has been masked, and the network needs to restore the pixel level semantic information in the patch based on the contextual information of the video.} 
\label{fig:masked}
\end{center}
\end{figure*}

Fig. \ref{fig:masked} showcases the segmentation results on the VSPW validation set using our trained model. The input images were randomly masked with $128 \times 128$ patches, and the visualizations display the model's ability to segment these masked images. Our model has learned to semantically restore the masked areas by leveraging both spatial and temporal contextual information. This additional training objective encourages the model to make more effective use of contextual cues, leading to significant improvements in segmentation performance on the original, unmasked images.

\begin{figure*}[!t]
\begin{center}
\includegraphics[width=1\linewidth]{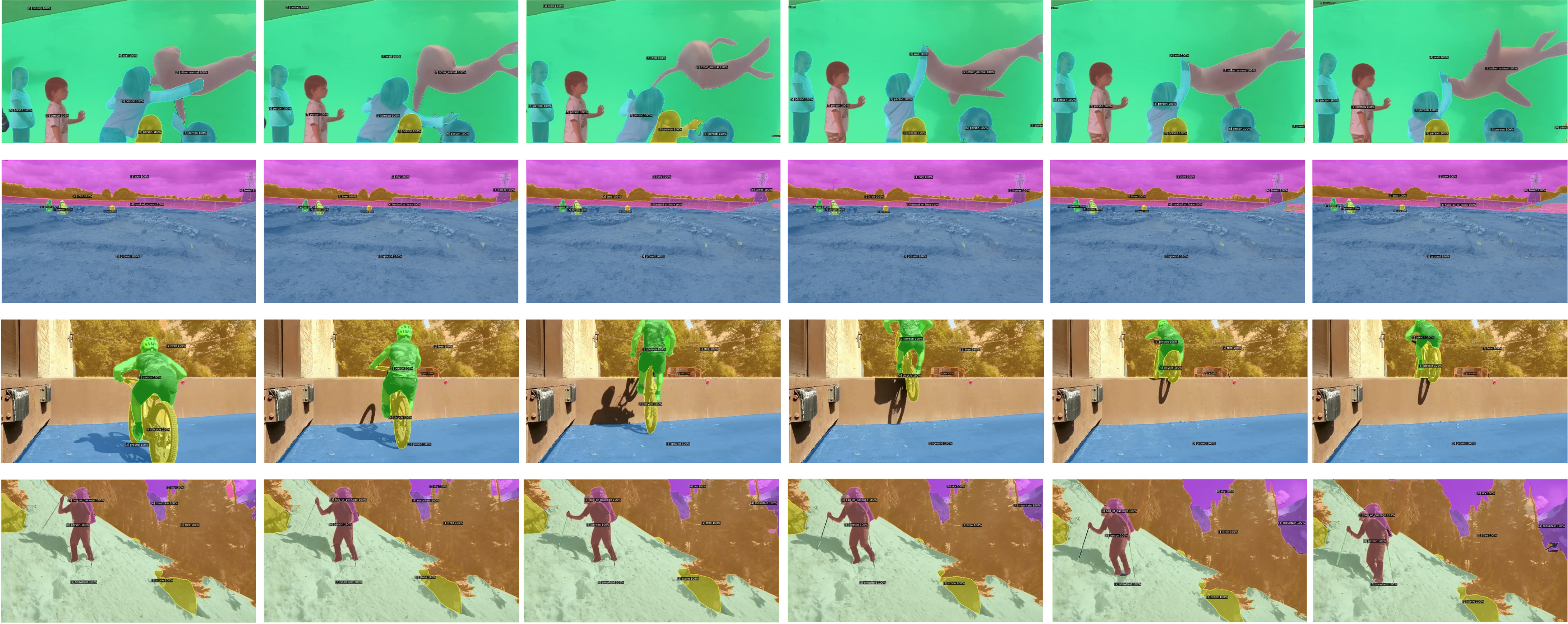}
\caption{Video panoptic segmentation results on VIPSeg} 
\label{fig:vipseg}
\end{center}
\end{figure*}

\begin{figure*}[!t]
\begin{center}
\includegraphics[width=1\linewidth]{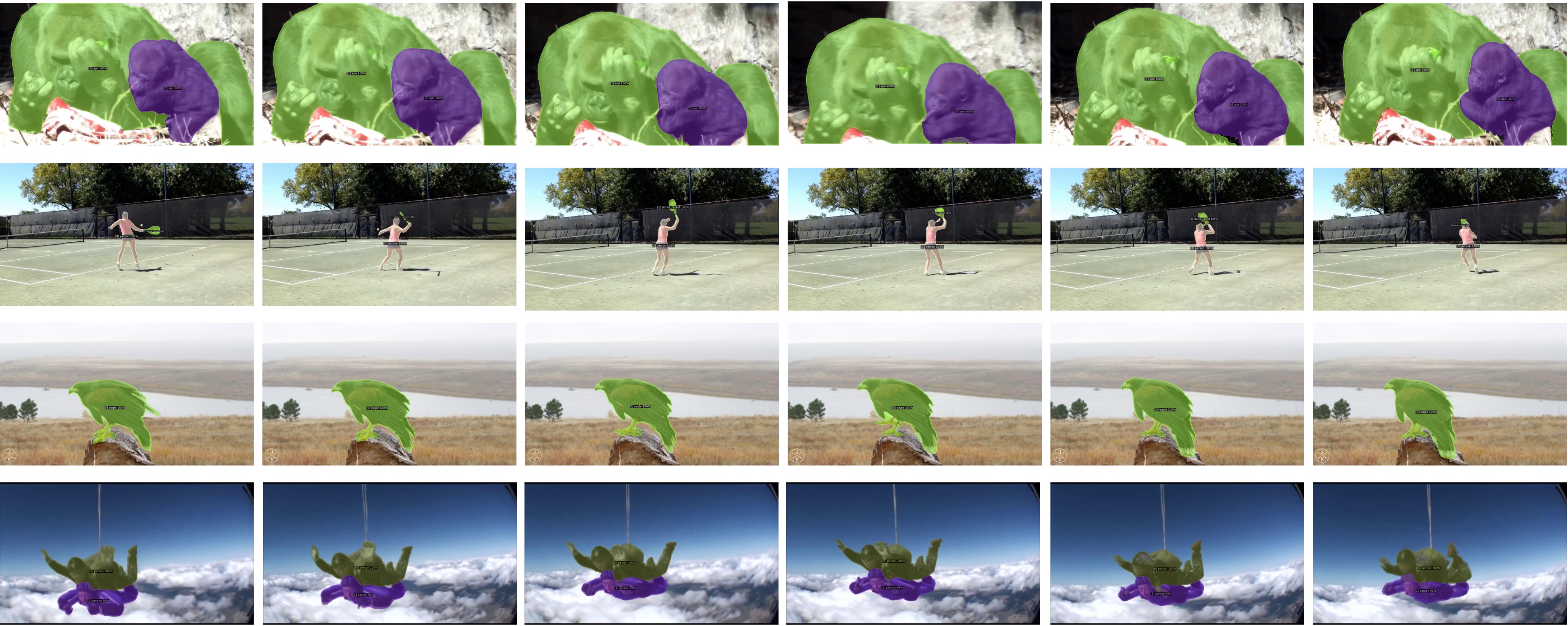}
\caption{Video instance segmentation results on OVIS} 
\label{fig:ovis}
\end{center}
\end{figure*}

\begin{figure*}[!t]
\begin{center}
\includegraphics[width=1\linewidth]{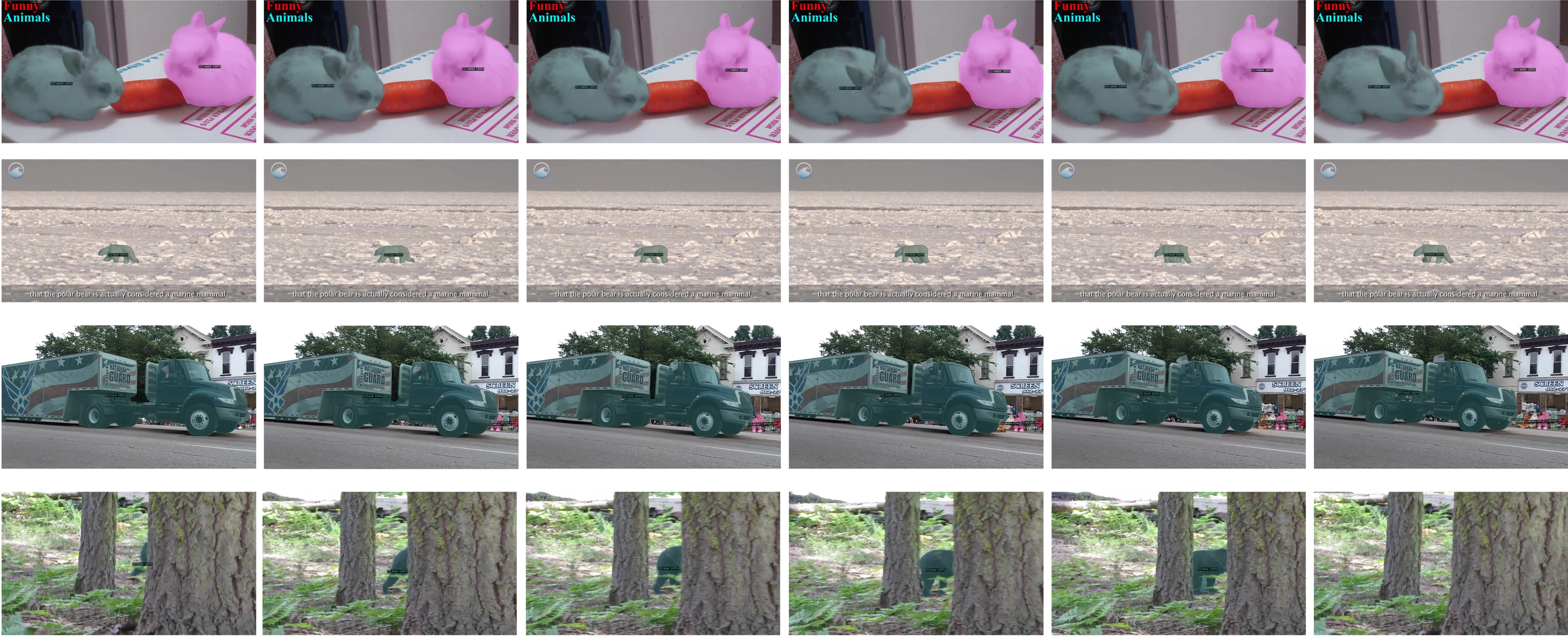}
\caption{Video instance segmentation results on YouTube VIS-2019} 
\label{fig:ytvis19}
\end{center}
\end{figure*}

\begin{figure*}[!t]
\begin{center}
\includegraphics[width=1\linewidth]{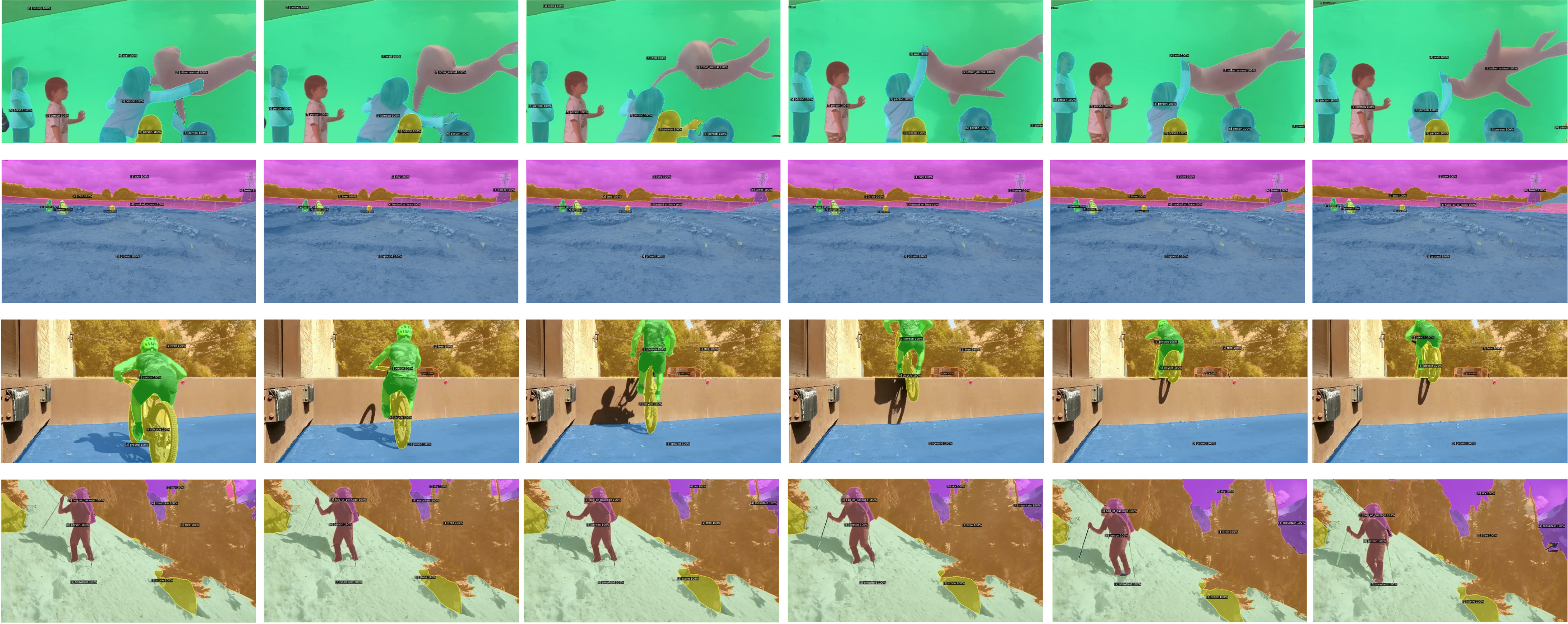}
\caption{Video instance segmentation results on YouTube VIS-2021} 
\label{fig:ytvis21}
\end{center}
\end{figure*}

Fig. \ref{fig:vipseg} - \ref{fig:ytvis21} illustrate the segmentation results of our method on video panoptic segmentation and video instance segmentation tasks. The results demonstrate that our approach enhances the segmentation quality for both "stuff" and "thing" categories. The segmentation masks of the same instance or semantic class exhibit relative continuity and consistency across different frames, and the model effectively focuses on objects of varying sizes. This indicates that even in complex scenes, the object queries extracted by the model maintain a high level of distinctiveness.

\end{document}